\documentclass[acmtog]{acmart}
\AtBeginDocument{%
  }

\copyrightyear{2026}
\acmYear{2026}
\setcopyright{cc}
\setcctype{by}
\acmConference[SIGGRAPH Conference Papers '26]{Special Interest Group on Computer Graphics and Interactive Techniques Conference Conference Papers}{July 19--23, 2026}{Los Angeles, CA, USA}
\acmBooktitle{Special Interest Group on Computer Graphics and Interactive Techniques Conference Conference Papers (SIGGRAPH Conference Papers '26), July 19--23, 2026, Los Angeles, CA, USA}
\acmDOI{10.1145/3799902.3811220}
\acmISBN{979-8-4007-2554-8/2026/07}

\acmSubmissionID{1506}

\usepackage{algpseudocode}
\usepackage{algorithm}
\usepackage{siunitx}
\usepackage{booktabs}
\usepackage{multirow}
\usepackage{svg}
\usepackage{subcaption}
\usepackage{microtype}
\usepackage{xspace}
\usepackage{enumitem}
\usepackage{algorithm}
\usepackage{algpseudocode}
\usepackage{amsmath}

\algnewcommand\algorithmicinput{\textbf{Input:}}
\algnewcommand\algorithmicoutput{\textbf{Output:}}
\algnewcommand\Input{\item[\algorithmicinput]}
\algnewcommand\Output{\item[\algorithmicoutput]}
\algrenewcommand{\algorithmiccomment}[1]{// #1}

\makeatletter

\algnewcommand{\LineComment}[1]{\Statex \hskip\ALG@thistlm // #1}

\DeclareRobustCommand\onedot{\futurelet\@let@token\@onedot}
\newcommand\@onedot{\ifx\@let@token.\else.\null\fi\xspace}

\newcommand{\eg}{\emph{e.g}\onedot} \def\Eg{\emph{E.g}\onedot}
\newcommand{\ie}{\emph{i.e}\onedot} 
 
\def\etc{\emph{etc}\onedot}

\makeatother

\NewDocumentCommand{\shiry}{ mO{} }{\textcolor{magenta}{\textsuperscript{\textit{Shiry}}\textsf{\textbf{\small[#1]}}}}
\NewDocumentCommand{\greg}{ mO{} }{\textcolor{green!50!black}{\textsuperscript{\textit{GS}}\textsf{\textbf{\small[#1]}}}}

\begin{document}

\title{Autoregressive Modeling of Film with Applications in Video Montage}

\author{Marcelo Sandoval-Casta\~neda}
\email{marcelo@ttic.edu}
\orcid{0000-0002-3227-9195}
\affiliation{%
  \institution{TTI-Chicago}
  \country{USA}
}

\author{Fabian Caba Heilbron}
\affiliation{%
 \institution{Adobe}
 \country{USA}
}
\email{caba@adobe.com}

\author{Shiry Ginosar}
\affiliation{%
 \institution{TTI-Chicago}
 \country{USA}
}
\email{shiry@ttic.edu}

\author{Bryan Russell}
\affiliation{%
  \institution{Adobe}
  \country{USA}
}
\email{brussell@adobe.com}

\author{Josef Sivic}
\affiliation{%
  \institution{Adobe}
  \country{USA}
  \and
  \institution{Czech Institute of Informatics, Robotics and Cybernetics, Czech Technical University}
  \country{Czech Republic}
}
\email{inr03127@adobe.com}

\author{Alexei Efros}
\affiliation{%
  \institution{UC Berkeley}
  \country{USA}
}
\email{aaefros@berkeley.edu}

\author{Gregory Shakhnarovich}
\affiliation{%
  \institution{TTI-Chicago}
  \country{USA}
}
\email{greg@ttic.edu}

\renewcommand{\shortauthors}{Sandoval{-}Casta\~neda et al.}

\begin{abstract}
This work introduces FilmGPT, an autoregressive transformer designed to address the challenge of {\em video montage} -- turning a collection of raw, ``unwatchable'' footage into coherent cinematic sequences.
Inspired by language learning in modern LLMs, we
train a long-context autoregressive transformer on a large corpus of movies.
The aim is to implicitly capture the ``grammar'' of film 
directly from data rather than from
hand-coded rules.
Unlike other generative models, FilmGPT does not generate any new video frames. Instead, at inference time, we introduce a footage-constrained decoding algorithm to
select
the best next shot from the input raw footage
according to the statistical patterns learned from films.
We first evaluate these learned statistics directly by using the FilmGPT autoregressive model for next shot prediction on a standard benchmark of shot sequence ordering, outperforming the previous state of the art. We then evaluate our footage-constrained decoding algorithm on the full film editing task via a user study,
and find that our FilmGPT-based editing
significantly outperforms previous approaches.
Finally, we demonstrate the applicability of FilmGPT to a wide range of applications in video montage, from automatic video segment trimming to human-in-the-loop film editing.
\textbf{Please see our supplementary material for qualitative results.}
\end{abstract}



\begin{teaserfigure}
    \begin{center}
        \href{https://mudtriangle.com/filmgpt}{\textcolor{blue}{\huge{\texttt{https://mudtriangle.com/filmgpt}}}}
    \end{center}
\end{teaserfigure}

\begin{teaserfigure}
    \centering
    \includegraphics[width=\textwidth]{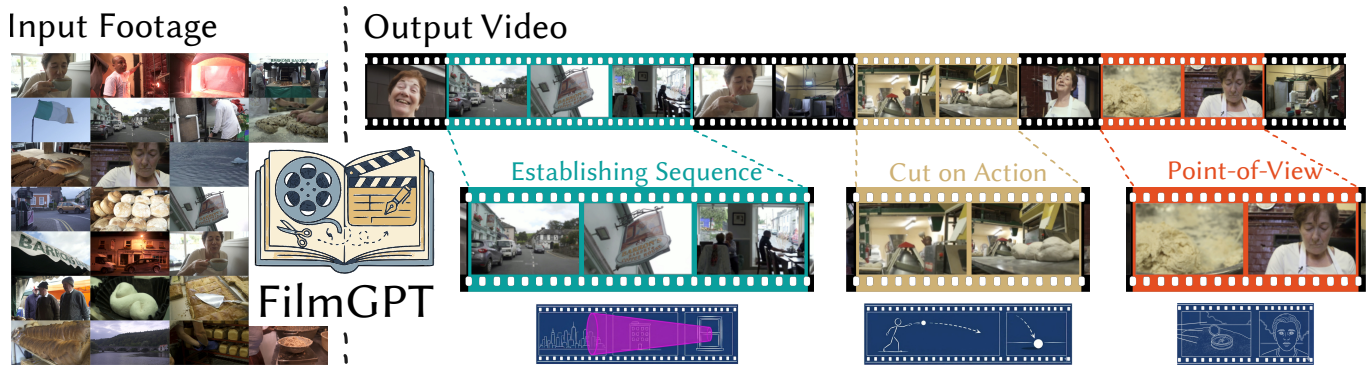}
    \vspace{-7mm}
    \caption{\textbf{From raw footage to edited video with emergent idioms.} Our system takes as input a collection of raw footage, typically hundreds of minutes. Our autoregressive model FilmGPT selects, trims and assembles videos from this collection into plausible sequences of shots. We find that this approach produces edited sequences where standard film editing idioms, such as establishing sequences, cutting on action, and point-of-view shots, naturally emerge. Thumbnails are copyrighted and belong to EditStock.}
    \label{fig:teaser}
\end{teaserfigure}

\maketitle

\section{Introduction}\label{sec:intro}

Video authoring and filmmaking once belonged to large crews, specialized equipment, and sky-high budgets. Over the last two decades, that barrier has collapsed. 
Digital cameras -- and then smartphones -- made capture cheap, portable, and nearly effortless.
Weddings, birthdays, school plays, camping trips, and everyday moments are now routinely recorded, often by several people at once. The result is an unprecedented abundance of raw video: hours, sometimes tens of hours, of footage for a single event.

And yet, most of this footage is never watched.
The reason is simple and well understood by filmmakers: raw video is largely unwatchable. Cameras run too long, frames ``wander'', and moments repeat. Editing -- in the sense of {\em video montage}, the selection, trimming, and ordering of shots rather than pixel-level modification -- is what turns this unruly material into something coherent and engaging. In professional filmmaking, editing routinely discards the vast majority of captured footage -- often on the order of a 1:100 shooting ratio -- to leave behind a carefully shaped sequence of shots. This observation is hardly new. Indeed, Orson Welles once said that ``it is the very eloquence of the cinema that is constructed in the editing room.''~\cite{welles58}

Importantly, editing is not an ad-hoc or purely intuitive process. It is a craft with a well-defined vocabulary and a learned ``grammar''. Film schools and textbooks devote extensive attention to film editing idioms, such as continuity editing, cutting on action, shot-reverse-shot patterns, match cuts, pacing, and visual rhythm~\cite{arijon1991grammar,katz1991film}. These conventions are taught because they work: when applied skillfully, they make edits feel “invisible”, preserving coherence and guiding the viewer without calling attention to the edit itself. At the same time, any experienced editor knows that these rules are flexible, frequently bent, and occasionally broken in service of the material. This combination -- structured principles coupled with contextual judgment -- makes editing both powerful and difficult to formalize.

It is useful to think of editing techniques as operating at multiple levels of abstraction. At the lowest level are shot-quality decisions: removing technically flawed material, trimming unusable segments, and ensuring that individual shots meet basic visual standards. At the mid level are structural decisions that make sequences coherent: continuity editing, cutting on action, respecting screen direction, assembling common patterns such as shot-reverse-shot or establishing sequences, and controlling pacing. At the highest level lie narrative and thematic decisions -- what a scene is about, what emotional arc it serves, and how sequences relate across an entire film. Our present work concentrates on the low- and mid-level abstractions. %

These abstractions are difficult to encode explicitly and to automate. The “grammar” of editing is not a set of rigid rules that can be cleanly programmed: conventions are often violated deliberately, context matters deeply, and acceptable edits are non-unique. The combinatorial space of possible edits grows explosively with footage length, making exhaustive search infeasible. Moreover, many mid-level techniques, such as dialogue shot-reverse-shot\footnote{A dialogue shot-reverse-shot is an editing technique that depicts a conversation between two characters by cutting back and forth between people speaking and reacting to each other.}, require reasoning over extended temporal windows, beyond individual frames or short clips. This reasoning is needed both to decide which shots to select and where within an action to place cuts, including identifying corresponding moments across shots. As a result, prior automated editing systems have largely relied on low-level signals such as motion or saliency, producing results that are highlight-driven but often brittle~\cite{gygli2016video2gif,xiong2019lessismore}.

Our approach takes a different path. We make the following technical contributions. %
(1)~\textbf{We frame the task of computational video montage as a next-token prediction problem over discrete video tokens.} We propose a long-context autoregressive transformer (dubbed {\em FilmGPT}, illustrated in Fig.~\ref{fig:teaser}), optionally conditioned on audio, that predicts sequences of tokens corresponding to frames drawn from a collection of raw footage. 
This formulation allows the model to learn the ``grammar'' of film editing directly from data rather than from hand-coded rules, while operating strictly on existing visual material.
(2)~\textbf{We make editing a first-class operation in video modeling} by introducing an explicit $\langle \text{CUT} \rangle$ token and a modified cross-entropy objective, enabling the model to jointly reason about shot selection and trimming.
(3)~\textbf{We design an inference-time footage-constrained decoding algorithm} that applies the learned model to assembly of raw footage into edited video sequences that follow standard cinematic editing conventions.
(4)~\textbf{We demonstrate the effectiveness of our approach through both qualitative and quantitative evaluation}, focusing primarily on professional-grade footage where editing conventions are clearly expressed and evaluation is more controlled. This includes a user study showing strong human preference over prior automated editing methods, as well as state-of-the-art performance on the AVE~\cite{argaw2022anatomy} shot sequence ordering benchmark. Finally, we demonstrate the model's versatility in professional applications such as multi-camera editing and stringout assembly.

\begin{figure*}[tbh!]
    \centering
    \includegraphics[width=\textwidth]{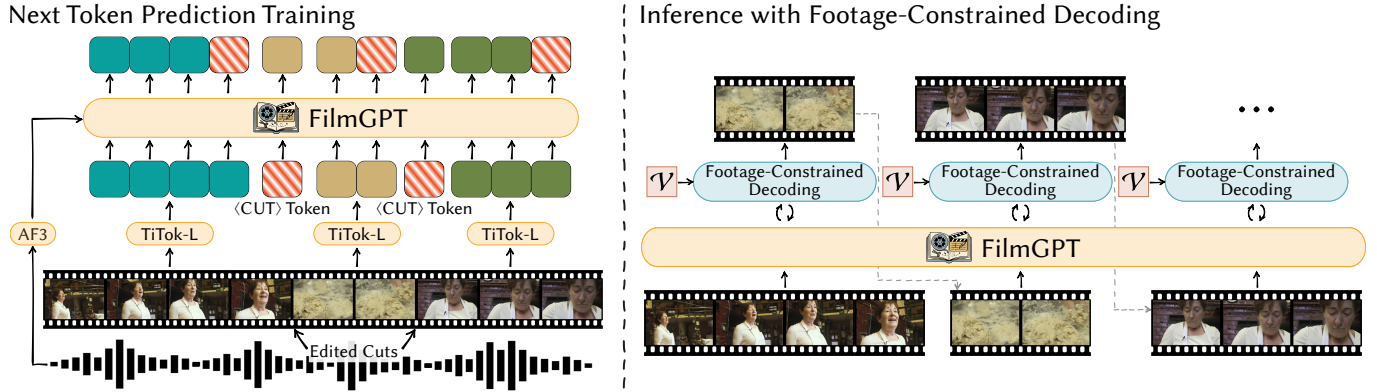}
    \caption{\textbf{Overview of our computational video montage framework.} ({\em Left}) During training, FilmGPT from movies via next token prediction. Given an edited video, each frame is individually tokenized using TiTok-L~\cite{yu2024an}, and an additional $\langle \text{CUT} \rangle$ token is added at the boundaries of each shot. For the purposes of this visualization, each frame is tokenized to one single token. In our actual implementation, each frame is tokenized to a sequence of $32$ tokens. Tokens belonging to the same shot are shown in the same color. The $\langle \text{CUT} \rangle$ token is shown with red diagonal stripes. FilmGPT is then trained as an autoregressive model using next token prediction with interleaved cross-attention layers that attend to the audio features produced by Audio Flamingo 3~\cite{goel2025audio}. ({\em Right}) At inference time, we produce edited video sequences using our footage-constrained decoding algorithm. At each step, the sampling algorithm chooses and trims a new shot from the video collection $\mathcal{V}$, which is then appended to the original context to select the shot that follows. This process is repeated until the desired number of shots (or duration) is reached. Thumbnails are copyrighted and belong to EditStock.}
    \label{fig:ar}
\end{figure*}

\section{Related Work}

\subsection{Computational Models of Film Editing Idioms}
Film idioms are the underlying structural units in which \emph{shots} (video sub-clips delimited by two cuts within a longer movie) are ordered to convey meaning~\cite{metz1992}.
There have been many attempts at understanding and applying movie idioms computationally.
Early works formalize idioms explicitly: hierarchical expert systems for movie editing~\cite{karp1993automated}, declarative programming languages for camera control~\cite{christianson1996declarative}, and finite state machines~\cite{he1996virtual}.
Further work decomposes the movie editing process into a series of actions to be performed iteratively using these formalizations~\cite{jhala2005discourse}.
The movie editing task can also be modeled as path finding in a graph where nodes are sub-clips, and the edges have costs obtained from hand-crafted functions based on idioms and other constraints~\cite{lino2011computational}.
Other approaches aimed to develop query and query-like languages with which to describe cinematographic properties of individual shots~\cite{ronfard2015prose}, stylistic patterns and constraints~\cite{wu2016analysing}, or both~\cite{wu2018thinking}.
The first approach to learn film editing patterns directly from data used a Hidden Markov Model (HMM) and relied on manual annotation of cinematographic data~\cite{merabti2016virtual}.
A competing approach showed that hand-crafting HMM parameters following a list of idioms was more effective in producing high-quality edits without costly annotations~\cite{leake2017computational}.

Most recently, \citet{Argaw_2024_CVPR} propose the Trailer Generation Transformer (TGT), an encoder-decoder transformer for sequencing shots into movie trailers.
TGT represents a movie as a sequence of shot embeddings (\eg, CLIP features) and learns, from curated movie-trailer pairs, to generate a trailer shot sequence autoregressively.
Our approach is related in spirit to TGT -- both cast editing as sequence modeling -- but differs in objective and supervision.
We model movies as sequences directly, as opposed to assembling trailers, and thus do not require aligned movie-trailer pairs.
Moreover, TGT operates on fixed-size shots, whereas our model operates at the frame level with an explicit cut operation, using a decoder-only autoregressive (AR) transformer and a discrete image tokenizer.
This is also reminiscent of~\citet{merabti2016virtual}, which requires manual annotations. In contrast, our supervision reduces to shot boundaries, which can be obtained reliably with off-the-shelf detectors such as TransNet V2~\cite{transnetv2}.

Finally, our formulation is also inspired by the classic Video Textures work~\cite{schodl2000video}.  Video Textures addresses the complementary problem of synthesizing arbitrarily long, visually coherent sequence from a short clip. It represents a video texture as a Markov chain in which each state is a frame and transition probabilities are derived from frame similarity. Then, synthesis proceeds by traversing this transition graph, selecting the next frame according to its transition probabilities. We adopt a similar sequential generation view, but for editing: FilmGPT predicts the next discrete visual token and uses an explicit cut token to decide when to cut.

\subsection{High-level Approaches for Computational Video Montage}
Many systems have been proposed to tackle specific sub-tasks in movie editing.
Some of these include predicting cuts in interviews~\cite{berthouzoz2012tools} to assemble a story, editing musical performances~\cite{lee2022popstage}, gaze-informed editing from a wide shot~\cite{moorthy2020gazed}, or determining transitions between clips that are fixed and given to the model~\cite{shen2022autotransition, guhan2025v}.
In addition, there are works that address the development of user interfaces that allow for interactive video editing using AI systems~\cite{huber2019b, huh2023avscript, tilekbay2024expressedit, wang2024lave, wang2024podreels}.

Other recent approaches leverage pre-trained language models to produce movies from existing footage.
The objective of these approaches is fine-grained user control by using natural language.
For example, Transcript2Video~\cite{xiong2022transcript} takes in a text query and retrieves footage that is most similar to the text query in a joint text-image embedding space.
TimelineAssembler~\cite{pardo2024generative} requires text prompts of explicit editing actions to execute, \eg, ``swap the order of the first two shots'', and fine-tunes a language model that outputs resulting edited videos.
Finally, EditDuet~\cite{sandoval2025editduet} models the video editing task as an interaction between two LLM-based agents that aim to satisfy a high-level user request expressed in natural language.

Our work targets a different level of abstraction than these language-driven editing systems. Transcript2Video, TimelineAssembler, and EditDuet focus on high-level intent and user control. They use language to decide what to include and how to change a timeline. Our FilmGPT does not reason over natural language. It models the low- and mid-level editing layer: selecting and trimming footage, placing cuts, and shaping local coherence from audiovisual cues. Nevertheless, in Sec.~\ref{sec:user-study}, we show that FilmGPT can be wrapped with a language interface to enable a direct comparison to prior language-driven editing systems. Given a user request, an LLM proposes a candidate set of shots from the footage, and FilmGPT turns that set into an edit by selecting, trimming, cutting, and pacing the final sequence. This isolates the low- and mid-level editing problem while keeping the narrative intent in the language space.

\section{Computational Video Montage Framework}
Our goal is to develop a system that outputs a video montage given a collection of raw footage. Such a system must make a series of editing decisions -- selecting shots, placing cuts to trim them, and ordering the resulting trimmed shots into a coherent sequence. Our key insight is that, while raw footage is unstructured, movies exhibit recurring patterns governing these decisions. These patterns correspond to common film idioms~\cite{metz1992,arijon1991grammar}. By training an autoregressive model on movies, we can learn these patterns from data and use it to assemble raw footage into a coherent video montage.

We formalize this idea by representing movies as sequences of frames and explicit cut events, encoded as discrete tokens interleaved with a special $\langle \text{CUT} \rangle$ token. Let $\mathbf{x}=(x_1,\dots,x_T)$ denote this sequence, with each $x_t$ drawn from a discrete vocabulary. In practice, each video frame is encoded as a short run of tokens (32 in our implementation), and we insert a single $\langle \text{CUT} \rangle$ token between shots.

We model the joint probability distribution $p(\mathbf{x})$ of professionally edited video sequences and factorize it autoregressively:
\begin{equation}
p(\mathbf{x}) = \prod_{t=1}^{T} p(x_t \mid x_1, \dots, x_{t-1}) .
\end{equation}
This formulation casts editing as causal next-token prediction over frames and cuts, in direct analogy to language modeling.

Unlike standard language generation, where token sequences can be arbitrary, an edit must be constrained to the footage collection. In our formulation, this means we cannot freely generate tokens from the vocabulary. Instead, we decode under a hard footage constraint: between $\langle \text{CUT} \rangle$ tokens, the visual tokens must correspond to a contiguous subsequence existing in the set of the raw input videos.

Our framework therefore consists of two main components, as shown in Fig.~\ref{fig:ar}:
\begin{enumerate}[leftmargin=*, label=\arabic*.]
\item An autoregressive model, FilmGPT, which learns a joint distribution $p(\mathbf{x})$ over sequences of frame and cut tokens from professional movies (Sec. ~\ref{sec:filmgpt}).
\item An inference-time footage-constrained decoding algorithm, which finds the highest-likelihood edit while enforcing that each shot is a contiguous segment of the footage collection (Sec. ~\ref{sec:constrained-decoding}).
\end{enumerate}

\subsection{FilmGPT: Autoregressive Modeling of Film}
\label{sec:filmgpt}
We now describe the instantiation of FilmGPT used to model movies.
Specifically, we define the discrete sequence $\mathbf{x}$ and parameterize $p(x_t \mid x_{<t})$ with a long-context autoregressive transformer, using a cut-weighted loss and optional audio conditioning.

\subsubsection{Discrete Tokenization of Edited Videos.}
To build FilmGPT, we represent a movie as a single discrete token sequence $\mathbf{x}=(x_1,\dots,x_T)$ that interleaves frame and cut tokens.
We use TiTok-L~\cite{yu2024an} to encode each frame into 32 discrete tokens from a 4096-token vocabulary.
We chose this method as it is a one-dimensional sequence of tokens, it produces short sequences compared to other discrete tokenizers, and it empirically yields sufficient reconstruction quality. For qualitative examples of TiTok-L reconstruction on our data, see the supplementary material.
Since our goal is to model filmmaking choices, we also represent each ``cut'' (\ie, the change from one continuous camera stream to the next within a long video) explicitly with a $\langle \text{CUT} \rangle$ token.
This makes cut placement a first-class event the model can learn to predict.
Finally, we prepend and append  beginning- and end-of-sequence tokens.

\subsubsection{Autoregressive Transformer with Long Context.}
Given the discrete token sequence $\mathbf{x}$ defined above, FilmGPT parameterizes the conditional distribution $p(x_t \mid x_{<t})$ with a decoder-only autoregressive transformer trained for next-token prediction.
As the above tokenization procedure yields an average length of 102k tokens for a single video in the AVE dataset~\cite{argaw2022anatomy}, 
naively applying full self-attention is expensive due to its quadratic complexity in sequence length. At the same time, editing decisions require both temporally fine-grained cues and long-range context, so it would be detrimental to truncate or subsample this context.

To make long-context modeling feasible, we adopt sliding window attention~\cite{Beltagy2020Longformer}, which reduces self-attention cost from $\mathcal{O}(n^2)$ to $\mathcal{O}(nw)$, where $n$ is the sequence length and $w$ the sliding-window size.
A window size of 16k tokens enables our model to train with a 128k-token context. Ignoring cut tokens, at 32 tokens per frame, this is roughly equivalent to 4000 frames or, at 24 frames per second, more than 2.5 minutes of video. In comparison, the average sequence of shots in the AVE dataset~\cite{argaw2022anatomy} is 2.25 minutes long, and instructional filmmaking books point at 3 minutes as the maximum suggested duration of a scene~\cite{trottier1994screenwriter}.
\textit{For more details on the model and training, see the supplementary material.}

\subsubsection{Cut-Weighted Cross Entropy Loss.}
The training sequences are long and contain few cuts:
on average, a 128k token sequence extracted from a typical movie scene contains only 34 $\langle \text{CUT} \rangle$ tokens.
As a result, training with standard cross-entropy loss encourages the model to spend most of its capacity on predicting the next token within a shot, rather than on learning when to cut and what shot should follow.
For our task, these cut decisions are a primary modeling target.
Thus, we modify the standard cross-entropy loss used in language modeling, introducing position-dependent weights that upweight cut tokens and the short sequence immediately after a cut.
Formally, $x_i$ denotes the ground-truth token at position $i$ and $\hat{p}_i$ the model's predicted distribution over the token vocabulary, such that:
\begin{align}
    \ell_i &= \delta_i \cdot CE(x_i, \hat{p}_i),\\
    \delta_i &= \begin{cases}
        w - d^{\text{cut}}_i, & \text{if } d^{\text{cut}}_i < w,\\
        1, & \text{otherwise,}
    \end{cases}
\end{align}
where $CE(\cdot,\cdot)$ is the standard cross-entropy loss, $w$ is the maximum weight given to a cut token, and $d^{\text{cut}}_i$ denotes the distance from position $i$ to the nearest preceding $\langle \text{CUT} \rangle$ or zero, if the token at position $i$ is itself a cut.
This ensures that the loss weight is larger when the target is a $\langle \text{CUT} \rangle$ token, and it quickly decays back to standard cross-entropy after a small window $w$ of tokens that follow.
We empirically set $w=32$, matching the number of tokens that correspond to a single frame.

\subsubsection{Audio Conditioning.}
Movie editing decisions are often driven by sound cues, especially for pacing and cut timing. 
Therefore, we additionally condition our model on audio features from the edited sequences during training.
We extract audio representations using Audio Flamingo 3~\cite{goel2025audio}, then mean pool these to one feature vector per second for efficiency. During training, we drop the audio features with probability $0.4$ to prevent our model from over-reliance on audio.

\begin{figure}
    \centering
    \includegraphics[width=\columnwidth]{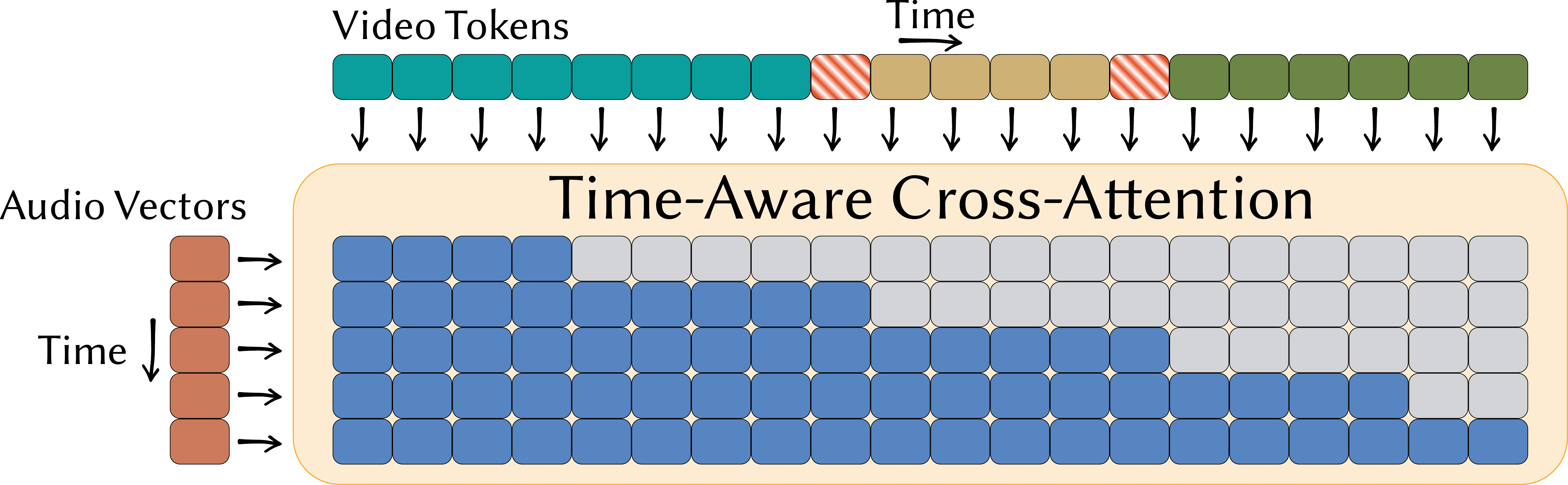}
    \caption{\textbf{Audio conditioning via cross-attention.} For the purposes of this visualization, each audio token corresponds to one second and each video token corresponds to $0.25$ seconds. Each video token only cross-attends to the audio vectors corresponding to the past and to the current second in time. The $\langle \text{CUT} \rangle$ token is shown with red diagonal stripes.}
    \label{fig:audio}
\end{figure}

The audio features are introduced to our autoregressive model by interleaving self-attention blocks with cross-attention blocks where the keys and values come from audio features~\cite{dong2018speech}.
As the audio for the next shot may inform what the current visual content should be, a given token in a sequence is only allowed to attend to audio features on the vector corresponding to the current second and to all vectors in the past, but not the future.
A cut always attends to the same audio vectors as the token immediately preceding it. Our audio conditioning is illustrated in Fig.~\ref{fig:audio}.

\begin{figure*}
    \centering
    \includegraphics[width=\textwidth]{fig/fig-userstudysample.pdf}
    \caption{\textbf{Samples from the three competing methods in our user study.} Transcript2Video~\cite{xiong2022transcript} often suffers from repeated shots. EditDuet~\cite{sandoval2025editduet} shows high bias towards shots of people talking, and shifts between characters doing seemingly unrelated tasks. In contrast, our method sticks to a specific set of characters, and shows clear progression in the task of baking bread across the entire sequence. See the supplementary material for these video outputs. Thumbnails are copyrighted and belong to EditStock.}
    \label{fig:user-study-sample}
\end{figure*}

\subsection{Footage-Constrained Decoding Algorithm}
\label{sec:constrained-decoding}

\begin{algorithm}[htbp]
\caption{Footage-Constrained Decoding Algorithm}
\label{alg:constrained_decoding}
\begin{algorithmic}[1]
\Input Model $\mathcal{M}$, Video collection $\mathcal{V}$, Init sequence $s$, Beam width $k$, Max shots $N_{\text{shots}}$, Max tokens $N_{\text{tokens}}$
\Output Edited token sequence $s$

\Function{GenerateVideo}{$\mathcal{M}, \mathcal{V}, s, k, N_{\text{shots}}, N_{\text{tokens}}$}
    \State $c_{\text{shots}} \gets 0$
    \While{$c_{\text{shots}} < N_{\text{shots}}$ \textbf{and} $\text{length}(s) < N_{\text{tokens}}$}
        \LineComment{\textbf{Step 1: Produce unique candidates}}
        \State $A \gets \text{ConstrainedBeamSearch}(\mathcal{M}, \mathcal{V}, s, k)$
        \State \textbf{if} $A = \emptyset$ \textbf{then break}
        
        \State $s_{\text{best}} \gets \text{null}$, $L_{\text{best}} \gets \infty$, $v_{\text{best}} \gets \text{null}$
        
        \LineComment{\textbf{Step 2: Find optimal length of each candidate}}
        \For{each video-position pair $(v, p) \in A$}
            \State $F \gets \text{ExtractTokens}(\mathcal{V}, v, p)$
            \State $s_{\text{cut}}, L_{\text{cut}} \gets \text{FindCut}(\mathcal{M}, s, F)$
            
            \LineComment{\textbf{Step 3: Finalizing the next shot}}
            \If{$L_{\text{cut}} < L_{\text{best}}$}
                \State $L_{\text{best}} \gets L_{\text{cut}}$,  $s_{\text{best}} \gets s_{\text{cut}}$, $v_{\text{best}} \gets v$
            \EndIf
        \EndFor
        
        \State \textbf{if} $s_{\text{best}} = \text{null}$ \textbf{then break}
        
        \State $s \gets \text{Concat}(s, s_{\text{best}})$ 
        \State $c_{\text{shots}} \gets c_{\text{shots}} + 1$
    \EndWhile
    \State \Return $s$
\EndFunction

\Statex
\Function{ConstrainedBeamSearch}{$\mathcal{M}, \mathcal{V}, s, k$}
    \State $B \gets \{s\}$ \Comment{Initialize beam}
    \While{\textbf{not} AllCandidatesUnique($B, \mathcal{V}$)}
        \State $B_{\text{next}} \gets \emptyset$
        \For{each candidate $c \in B$}
            \State $P \gets \text{NextTokenProbabilities}(\mathcal{M}, c)$
            \State $T_{\text{valid}} \gets \text{GetValidNextTokens}(\mathcal{V}, c)$
            \State $P \gets \text{MaskInvalidTokens}(P, T_{\text{valid}})$
            \State $T_{\text{top\_k}} \gets \text{GetTopK}(P, k)$
            \For{each token $t \in T_{\text{top\_k}}$}
                \State $B_{\text{next}} \gets B_{\text{next}} \cup \{\text{Concat}(c, t)\}$
            \EndFor
        \EndFor
        \State $B \gets \text{PruneToTopK}(B_{\text{next}}, k)$
    \EndWhile
    \State \Return \text{GetVideoPositions}($\mathcal{V}, B)$
\EndFunction

\Statex
\Function{FindCut}{$\mathcal{M}, s_{\text{base}}, F$}
    \State $I_{\text{valid}} \gets \text{TokensPerFrame}(F)$ \Comment{32 tokens in TiTok-L}
    \LineComment{Compute per-token NLL for all valid prefixes}
    \State $L_{\text{array}} \gets \text{PrefixCutNLL}(\mathcal{M}, s_{\text{base}}, F, I_{\text{valid}})$
    
    \State $i_{\text{opt}} \gets \arg\min_{i \in I_{\text{valid}}} L_{\text{array}}[i]$
    \State $s_{\text{best\_cut}} \gets \text{Concat}(F[0:i_{\text{opt}}], \langle \text{CUT} \rangle)$
    
    \State \Return $s_{\text{best\_cut}}, L_{\text{array}}[i_{\text{opt}}]$
\EndFunction

\end{algorithmic}
\end{algorithm}

Given our trained autoregressive model, we seek to edit raw footage from a video collection of any duration.
This task is challenging as there are combinatorial possible ways to select and trim footage from the collection.
We are inspired by constrained generation approaches that output a sequence of tokens one at a time using beam search~\cite{hokamp2017lexically}.
However, these works usually sample from the model following permissive sets of rules, whereas we need to ensure that the selected subsequences are constrained to the video collection, and that selected shots are separated with $\langle \text{CUT} \rangle$ tokens.
We address this by using a footage-constrained generation approach.
Let $\mathcal{V}$ be a tokenized video collection of raw footage, composed of videos of variable length.
We perform beam search over the space of possible edited videos assembled from subsequences in collection $\mathcal{V}$.

We start with a sequence of tokens $s$ to be extended. 
At a given step in the editing selection process, we maintain a beam of $k$ candidate extensions of a given length, say $m$ tokens. Critically, we enforce a hard subsequence constraint: every generated token prefix must match a contiguous span of tokens in at least one video in $\mathcal{V}$. The algorithm proceeds along the following three steps (details are in the supplementary material).

\subsubsection{Step 1: produce unique candidates.} We start with empty candidate sequences and extend them one token at a time, each time retaining the $k$ top valid candidates under the subsequence constraint. Since each frame is represented by 32 discrete tokens, short prefixes (\eg, the first 1--3 tokens) may match multiple frames in $\mathcal{V}$.
Thus, we keep extending each candidate until it has a unique anchor in $\mathcal{V}$, \ie, it matches a single frame occurrence (at a specific video and position). In practice, this requires fewer than 32 tokens.

\subsubsection{Step 2: finding optimal length of each candidate.} We extend each candidate by adding varying numbers of frames following it in the associated video in collection $\mathcal{V}$, up to 10 seconds, and select the extension which, appended with the cut token, has the highest per-token likelihood under the model. This can be done efficiently in FilmGPT, computing likelihoods in parallel for all possible lengths.

\subsubsection{Step 3: finalizing the next shot.} Step 2 yields $k$ candidate shots extending sequence $s$, potentially of different lengths. We sample one of these with weights proportional to their per-token likelihoods, or (if doing greedy decoding) select the one with the highest likelihood. In our experiments, we use greedy decoding.

Note that one can further restrict the process by, \eg, removing frame sequences already in sequence $s$ from the video collection $\mathcal{V}$.
We do not do this in our experiments, and observe that the model avoids such duplicate selections naturally.

Each round of this process yields the new extended sequence $s$, which is then fed to the next round as input to the model to be extended further, as shown in Fig.~\ref{fig:ar} (right). This results in our model producing a full edited sequence. %
There are multiple possible stopping conditions for this algorithm, depending on the application: desired number of shots in sequence, desired duration of sequence, number of distinct source videos selected from $\mathcal{V}$, \etc. The full pseudo-code is in Algorithm~\ref{alg:constrained_decoding}.

\section{Results}

We train FilmGPT using edited video sequences from several sources: AVE dataset~\cite{argaw2022anatomy}, movies in the public domain in the US, movies from the Soviet studio Mosfilm available online, and feature-length documentaries with permissive Creative Commons licenses, for a total of 6200 hours of training footage. The AVE dataset has scenes segmented into shots. For the other movies, we use an off-the-shelf shot change detector~\cite{transnetv2}.

\subsection{User Study: B-Roll Documentary Editing}\label{sec:user-study}

We ran a user study using the EditStock~\cite{editstock} evaluation setup proposed in EditDuet~\cite{sandoval2025editduet}.
It is a forced choice preference assessment where we ask participants to compare editing quality between samples from our model against baselines using the same raw footage.
The samples come from documentaries where each model is tasked with assembling B-roll sequences starting from a fixed A-roll sequence.
A-roll refers to voice-over or talking head interview footage, and B-roll refers to the complementary visuals that aid in storytelling.
This is a standard task in documentary filmmaking, where editors assemble A-roll as an initial step and then work on the corresponding B-roll.

Following EditDuet, we show participants the two videos with their corresponding audio tracks and ask them ``Which of the two video sequences is better edited?''
to assess general preference.
We compare results from our algorithm with results from Transcript2Video~\cite{xiong2022transcript} and EditDuet~\cite{sandoval2025editduet}.
We ran this study with a total of 83 participants on the Prolific platform, for a total of 1242 data points.
Fig.~\ref{fig:user-study-sample} shows one sample edited video for each method in our user study.

As our model does not support text instructions natively, we adapt our footage-constrained decoding algorithm (Sec.~\ref{sec:constrained-decoding}) and include a large language model (LLM) planner. We use the same LLM as in EditDuet (\texttt{Llama-3.1-8B-Instruct} \cite{dubey2024llama3herdmodels}).
The LLM receives a JSON file containing high level descriptions of every video in the raw footage and an instruction in plain text from the EditDuet user study.
Instructions are high-level descriptions of B-roll sequences, along with a desired duration.
The LLM then decomposes the instruction into descriptions of the visual components of each B-roll sequence, \eg, ``Tight close-ups of an old woman kneading dough, interleaved with close-ups of her face while baking. End with fast-paced shots of the oven setup process'' becomes ``close-ups of hands'', ``old woman face'', ``old woman opening oven'', \etc.
We then constrain our video collection to files that match these descriptions by computing CLIP~\cite{radford2021learning} similarity with key frames from the raw footage.
The top 50 video matches become the reduced video collection our model will sample from during editing.
This process is the only form of external conditioning that our model receives when assembling these sequences.

\begin{figure}[t]
    \centering
    \includegraphics[width=\columnwidth]{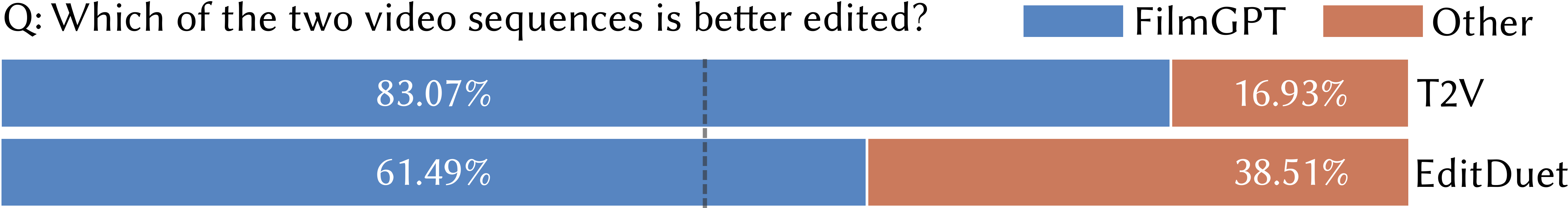}
    \caption{\textbf{Participant preference results in our user study.} We compare our model with T2V~\cite{xiong2022transcript} and EditDuet~\cite{sandoval2025editduet}. $p < 0.001$ for all results. Our method outperforms competing methods in terms of participant preference.
    }
    \label{fig:user-study}
\end{figure}

As shown in Fig.~\ref{fig:user-study}, our method is preferred by participants over the two methods we compare against.
Compared to Transcript2Video, users prefer our method $83.07\%$ of the time.
Compared to EditDuet, our method is preferred $61.49\%$ of the time.
Despite relying on the same LLM as EditDuet for story generation, our method's edits are favored over EditDuet's, without multiple rounds of LLM calls, iterations, or exploration.
In addition, EditDuet relies on pre-segmenting videos using TW-FINCH~\cite{tw-finch}, whereas our method can perform this task implicitly.

\subsection{Ablation Study: Shot Sequence Ordering}

We use the standard AVE~\cite{argaw2022anatomy} shot sequence ordering task as an indicator of performance when conducting ablations.
The task is as follows: given a set of three consecutive shots $s_0$, $s_1$, and $s_2$ from a scene, the model should determine the correct ordering of $s_0$, $s_1$, and $s_2$.
With our autoregressive model, we calculate the probability of each of the six possible orderings, and return the highest probability ordering.

Table~\ref{tab:accuracy} shows results of our ablations on shot sequence ordering.
We start from a base autoregressive (AR) transformer without $\langle \text{CUT} \rangle$ tokens and with a maximum context length of 16k tokens using full attention and no audio conditioning.
We train it on AVE only, and get an accuracy of $25.3\%$.
Explicitly representing each cut with a single token yields a significant improvement in accuracy, $29.7\%$.
Next, training the model with our cut-weighted cross-entropy loss, which forces the model to prioritize cuts over other portions of the sequence, improves accuracy up to $34.5\%$.
For comparison, we evaluate an alternative objective computed only at the final token of each frame and at $\langle \text{CUT} \rangle$ tokens; however, this yielded an accuracy of $19.2\%$, so we discard it.
The following step is to use sliding window attention in our autoregressive transformer so that its context can be extended up to 128k tokens.
At this step, we also start incorporating into the input the context of the sequence that we are trying to order.
For example, a video $s$ is composed of $t$ shots $s_0, s_1, \ldots, s_{t - 2}, s_{t - 1}$.
The task is to order the shots $s_{t - 3}, s_{t - 2}, s_{t - 1}$.
Our model can now receive $s_0, s_1, \ldots, s_{t - 4}$ as context to perform this task.
This is both a closer setup to the real task of editing and a way of including extra information for the task that previous approaches are unable to incorporate.
Accuracy with these additions jumps to $44.5\%$, showing how important it is for editing to have access to a larger context.
Adding audio conditioning also improves accuracy considerably, to $50.5\%$.
Audio from the context is included as-is, and audio from the sequence to be ordered is permuted to align with the six possible video orderings.
Finally, we add our large collection of movies as training on top of AVE.
This pushes accuracy to a final number of $53.9\%$, a full $18.6$ points of improvement over the previous state of the art method, UQNet~\cite{li2025shot}.
Note that UQNet leverages extra manual labels (shot size, camera angle, motion, genre). Despite using no such annotations, our model learns directly from the data and uses preceding context to implicitly capture rhythm, pacing, and cinematic language.

Our ablations are all trained with a fixed compute budget. Training of the base AR transformer, the addition of a $\langle \text{CUT} \rangle$ token, the addition of cut-weighted cross entropy loss, and the 128k context each take 131.4 EFLOPs, and inference cost in these models is 0.07 GFLOPs per token. Our audio conditioned model and extra data take 239.7 EFLOPs to train and 0.13 GFLOPs per token.

\begin{table}[tb!]
\centering

\caption{\textbf{Ablations on shot sequence ordering.} Previous state-of-the-art performance (UQNet~\cite{li2025shot}) included for reference. Best performance in \textbf{bold}. ``Acc@1'' refers to standard accuracy and ``Acc@3'' (first reported in UQNet), refers to the likelihood that the correct ordering is included within the top 3 predictions of the model.}
\label{tab:accuracy}

\begin{tabular}{l S[table-format=2.1] S[table-format=3.1] S[table-format=2.1] S[table-format=3.1]}
\toprule
\multicolumn{1}{c}{\textbf{Model}} & \textbf{Acc@1} & \textbf{$\Delta$} & \textbf{Acc@3} & \textbf{$\Delta$} \\
\midrule
UQNet~\cite{li2025shot} & 35.3 & $\text{---}$ & 69.8 & $\text{---}$ \\
\midrule
Base AR Transformer & 25.3 & $\text{---}$ & 57.1 & $\text{---}$ \\
\hspace*{4pt} + $\langle \text{CUT} \rangle$ Token & 29.7 & $\text{+}$4.4 & 67.3 & $\text{+}$10.2 \\
\hspace*{4pt} + Cut-Weighted CE & 34.5 & $\text{+}$9.2 & 68.9 & $\text{+}$11.8 \\
\hspace*{4pt} + 128k Context & 44.5 & $\text{+}$19.2 & 81.4 & $\text{+}$24.0 \\
\hspace*{4pt} + Audio Conditioning  & 50.5 & $\text{+}$25.2 & 84.1 & $\text{+}$ 26.7 \\
\textbf{\hspace*{4pt} + Our Datasets} & \textbf{53.9} & $\text{+}$28.6 & \textbf{84.3} & $\text{+}$26.9 \\
\bottomrule
\end{tabular}

\end{table}

\section{Applications in Video Montage}\label{sec:apps}
Our model enables a range of real-world video editing applications. We use FilmGPT as-is and apply the same constrained editing algorithm across tasks, with only minor constraints tweaks.
We present four applications, from low-level to mid-level complexity.

\begin{figure}[t]
    \centering
    \begin{subfigure}{\columnwidth}
        \includegraphics[width=\columnwidth]{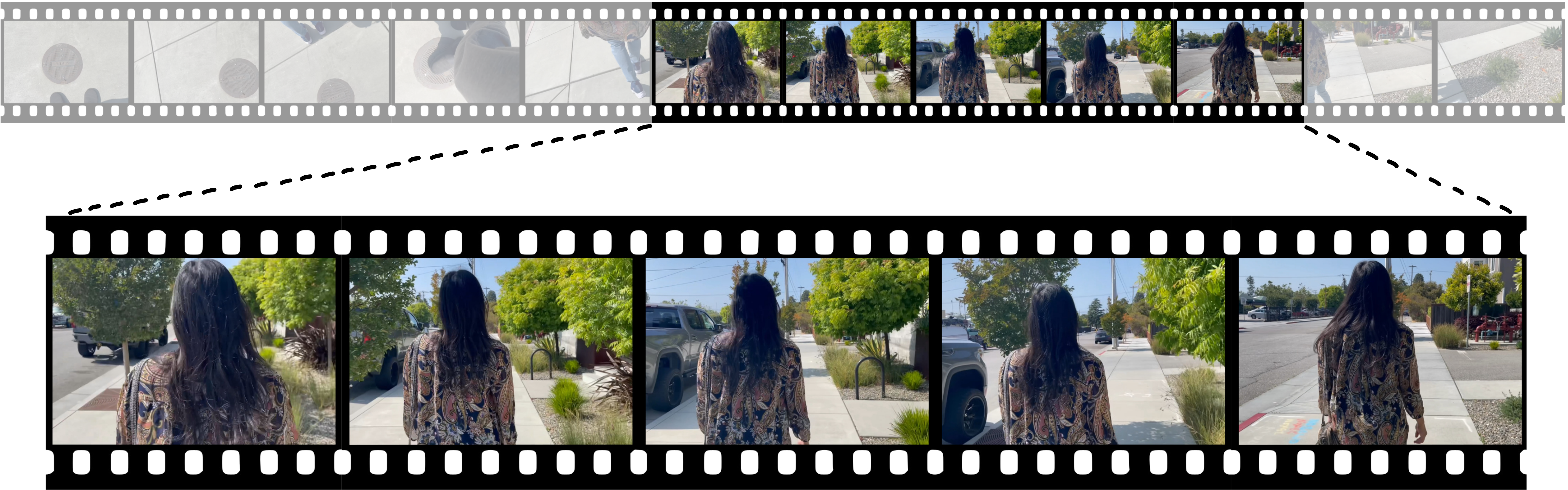}
        \caption{\textbf{Medium tracking shot from behind.} Our method is able to discard the first start of the video, where the camera is being set up while pointing at the floor and moving around erratically. It also discards the end of the video, where the camera quickly moves around as it is being turned off.}
        \label{fig:useful-segment-a}
    \end{subfigure}

    \bigskip

    \begin{subfigure}{\columnwidth}
        \includegraphics[width=\columnwidth]{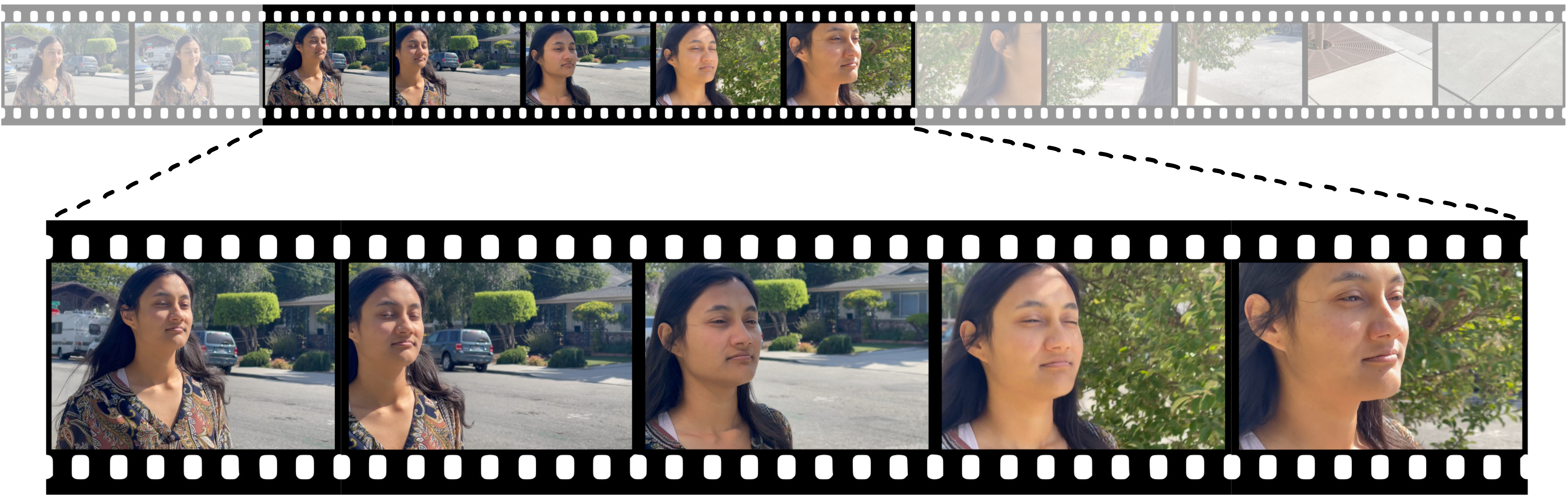}
        \caption{\textbf{Frontal tracking close-up shot.} Our method discards the first few frames where little is happening, and starts the sequence when the woman starts walking. Our method cuts right before the woman walks out of the frame.}
        \label{fig:useful-segment-b}
    \end{subfigure}
    \caption{\textbf{Samples from our video segment trimming example application.} Grayed out portions on the film strips are segments not selected by our method. The model correctly assigns low probability to segments that do not look cinematic and with erratic movements. Output trimmed videos are provided in the supplementary material.}
    \label{fig:useful-segment}
\end{figure}

\subsection{Video Segment Trimming}
When shooting film, it is common for videos to start and end with footage that is not meant to be used in the final product.
\Eg, in scripted fiction film this includes elements of the filmmaking process such as slates, cameras moving into and out of position, and other general setup of a scene.
A low-level task in video editing is to find the segments in each video that do not include such elements.

Our editing algorithm can be directly applied to this task.
The ``video collection'' in this application is the single video at hand.
We want the model to identify a contiguous sequence of tokens from the video, appended with the $\langle \text{CUT} \rangle$ token, that has the highest likelihood.
As our model is trained on professional movies, a sequence starting too early/late will yield a worse likelihood than trimming to the useful segment.

Fig.~\ref{fig:useful-segment} and the supplementary material provide two examples of this process with our own footage.
We use tracking shots where portions of the beginning and end consist of erratic camera movements and visual content of little interest, such as the ground and the shoes of the cameraperson. These portions would be extremely unlikely occurrences in a real movie and, in both examples, our method successfully trims them out.

\subsection{Assembling Stringouts}
Stringouts are defined as smaller sequences of shots that are coherent next to each other in a larger film. These are typically  sequences of two to four shots that follow certain patterns of editing, \eg, cutting on action or establishing sequences.

Stringouts are commonly assembled as the first step in the editing process. They are then used as building blocks to draw from, instead of drawing from raw footage directly, and then further refined by editors as needed. 

Our method is capable of coming up with stringouts given a video collection and an empty initial sequence. %
We employ our editing algorithm as-is over the entire collection of raw footage, and simply let it draw video sequences from any given file.
As there is no existing sequence that provides conditioning signal for pace or content, our method simply defaults to producing high likelihood sequences, which generally respect film editing conventions.
Though this property also arises in other editing tasks, we find that assembling stringouts in this way considerably increases the frequency and diversity of identifiable film idioms.

\begin{figure}
    \centering
    \begin{subfigure}{\columnwidth}
        \includegraphics[width=\columnwidth]{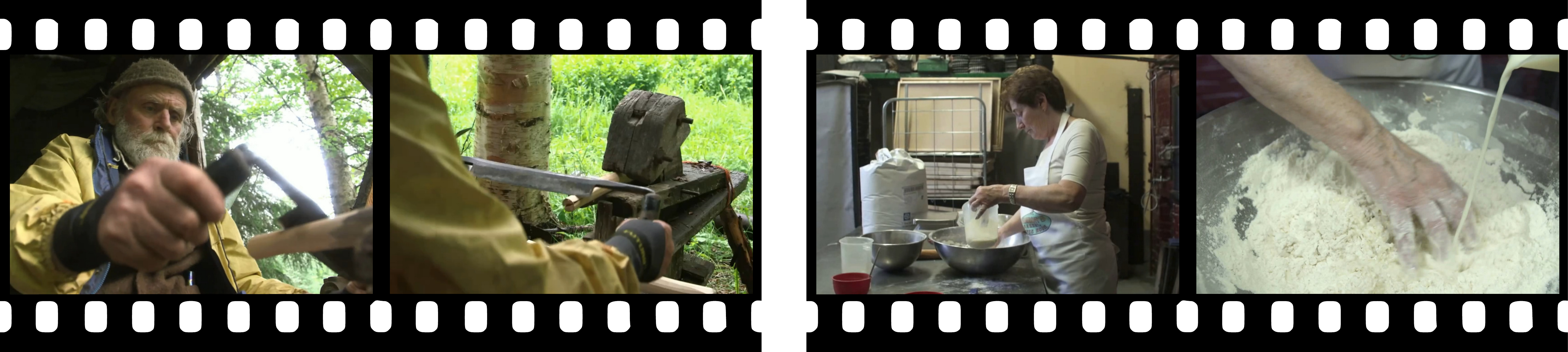}
        \caption{\textbf{Cut on action}. Consecutive frames from sequences assembled by our method for ``Built by Life'' and ``The Ovens of Cappoquin''. Note how the continuity of action is preserved across cuts in both sequences, despite the shots not being recorded at the same time.}
        \label{fig:stringouts-a}
    \end{subfigure}

    \bigskip

    \begin{subfigure}{\columnwidth}
        \includegraphics[width=\columnwidth]{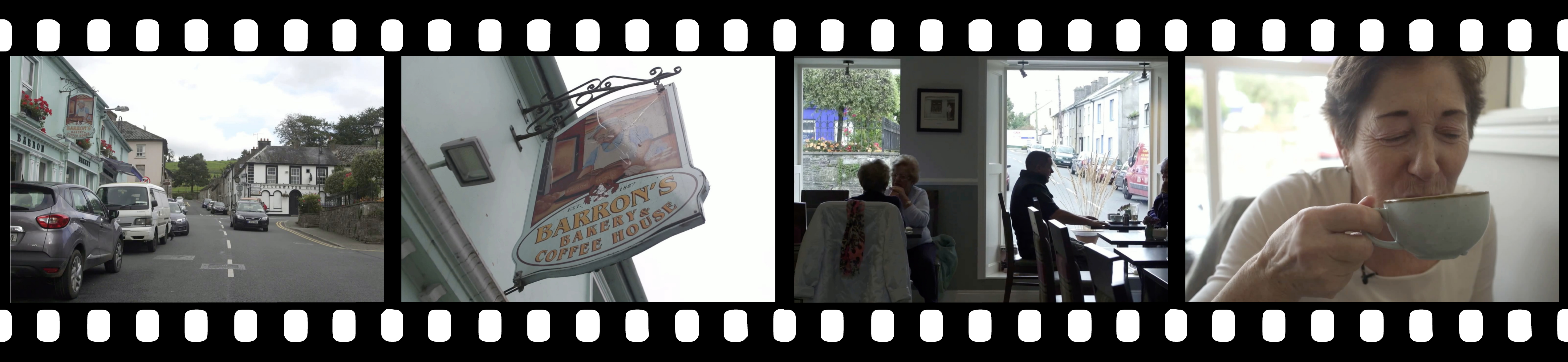}
        \caption{\textbf{Establishing sequence}. Sequence of shots assembled by our method for ``The Ovens of Cappoquin''. The video gets progressively closer to the protagonist: first, with a long shot outside the bakery, then with the bakery's street sign, followed by moving inside the bakery, and finally our protagonist's face in a close-up as she drinks tea.}
        \label{fig:stringouts-b}
    \end{subfigure}

    \bigskip

    \begin{subfigure}{\columnwidth}
        \includegraphics[width=\columnwidth]{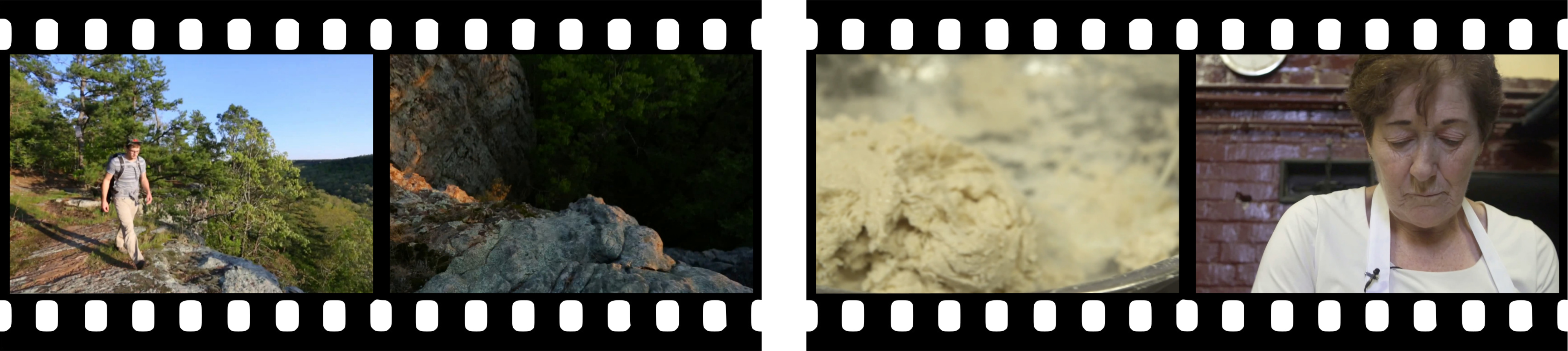}
        \caption{\textbf{Point-of-view}. Consecutive frames from sequences assembled by our method for ``The Rock Climber'' and ``The Ovens of Cappoquin''. The juxtaposition of these shots give the sensation that we see the objects (\ie, the cliff on the left sequence and the dough on the right sequence) from the perspective of the protagonists.}
        \label{fig:stringouts-b}
    \end{subfigure}
    \caption{\textbf{Samples from our stringouts example application.} Our method convincingly assembles short sequences of shots that respect simple movie idioms. Output stringout videos are provided in the supplementary material. Thumbnails are copyrighted and belong to EditStock.}
    \label{fig:stringouts}
\end{figure}

Fig.~\ref{fig:stringouts} and the supplementary material provide examples using footage from EditStock documentaries.
The input for the examples is the raw footage for each project (hundreds of minutes at $23.976$ frames per second).
Our method is able to assemble plausible stringouts based on film idioms -- cutting on action, establishing sequences, and point-of-view shots, which are widely used building blocks in film.

\begin{figure*}
    \centering
    \includegraphics[width=\textwidth]{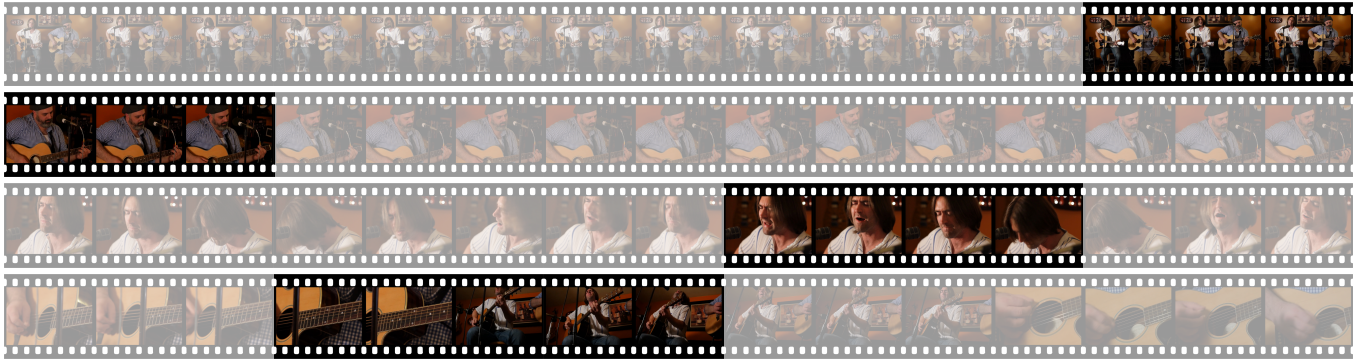}
    \caption{\textbf{Sample from our multi-camera editing example application.} Grayed out portions on the film strips are segments not selected by our method. Footage comes from the project ``Before You Accuse Me'', where a duo of musicians with guitars play in front of a number of cameras. Our method cuts between streams focusing on sources of audio and rhythm. The output edited video is in the supplementary material. Thumbnails are copyrighted and belong to CineStudy.}
    \label{fig:multicam}
\end{figure*}

\begin{figure*}
    \centering
    \includegraphics[width=\textwidth]{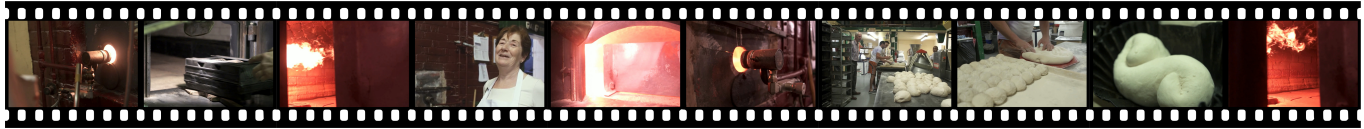}
    \caption{\textbf{Sample from our human-in-the-loop video editing example application.} The fourth frame in the strip belongs to the A-roll. Note the ability of our method to stick to thematic cross-cutting across B-roll sequences, where shots of the oven fire are interleaved with the process of preparing dough. See the supplementary material for the video outputs. Thumbnails are copyrighted and belong to EditStock.}
    \label{fig:hitl}
\end{figure*}

\subsection{Aligned Multi-Camera Editing}
Mutli-camera setups are common when shooting live music performances, theatre, and other live events that cannot afford multiple takes.
The cameras are placed in different locations in space and synchronized by external devices to maximize coverage of the event.
Our footage-constrained decoding can be directly applied to multi-camera raw footage, with one new constraint: the model can no longer draw from any frame in any file, but only those that correspond to the current time step in the video collection.
The editing task then becomes simply picking a frame from one of the video streams at a time and deciding when to change video streams.

Fig.~\ref{fig:multicam} and the supplementary material show an example output from our method.
For this, we use the project ``Before You Accuse Me'', from the film editing website CineStudy.
In the footage, a duo of musicians with guitars play a song in a small room facing toward the cameras.
The project consists of four video streams in sync: the first three are static cameras on a tripod and the fourth is a collection of aligned B-roll stringouts from multiple handheld cameras.
At every frame, our model selects whether to continue on the current video stream or to cut and switch to one of the other three.
Our method cuts appropriately to the pacing of the song and selects portions of the footage with the appropriate main subject.
It focuses on the correct guitars during guitar solos, moves to close-ups and medium shots of the person singing, and selects wider shots during the more uneventful portions of the music video.
In the segment shown in Fig.~\ref{fig:multicam}, our method maintains an editing pattern of medium shot -- close-up -- medium shot -- close-up during one of the guitar solos that then moves into singing, and breaks it when that particular section of the song is over.

\begin{figure}
    \centering
    \includegraphics[width=\columnwidth]{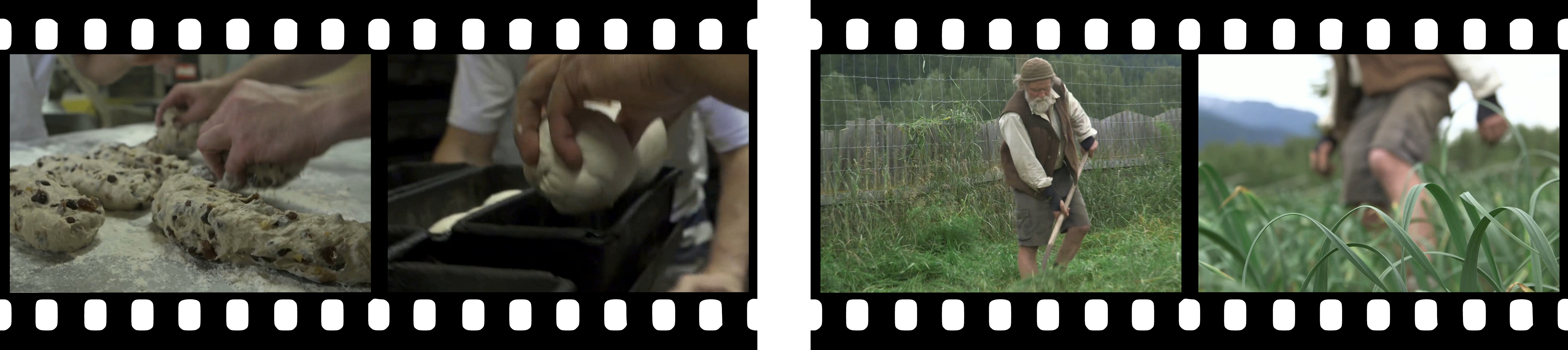}
    \caption{\textbf{Failure mode: cutting on unrelated actions.} Consecutive frames from sequences assembled by our method for ``The Ovens of Cappoquin'' and ``Built by Life''. On the left, a shot of hands grabbing dough is cut to a shot with hands placing dough in a loaf pan. The dough, the location, and the lighting are clearly different, making a jarring cut. On the right, a man working the field with a rake is cut to a man walking. Not only is the action different, but this cut also violates the 180-degree rule of editing, making a jarring cut. Thumbnails are copyrighted and belong to EditStock.
    }
    \label{fig:bad-cut}
\end{figure}

\subsection{Human-in-the-Loop Video Editing}
Finally, we show that our approach is well-suited for human-in-the-loop video editing.
In this application, the task is to assist editors in completing a full edited video.
We set this up by replacing the LLM in our user study with a human editor that is assembling an edited draft for a short film.
As the user edits, our method produces candidate sequences to continue the edit, and the human editor is able to choose and modify sequences produced by our method to edit instead of manually searching through the hundreds of minutes of raw footage.
For this use case, we maintain the conditioning by text description, and include further constraint options for the video collection based on the video metadata. These include constraining by file name, location, date, and file directory.
We include an example output of this process for ``The Ovens of Cappoquin'' in Fig.~\ref{fig:hitl} and in our supplementary material.
Producing an acceptable edited short film of 3.5 minutes from 288 minutes of raw footage took an amateur editor a few hours, as opposed to the several days the task often takes~\cite{larosa2025}.

\section{Limitations and Ethical Considerations}
Our approach has the following limitations. First, FilmGPT is limited by the context length. We believe that switching from the TiTok tokenizer to one built specifically for video requiring fewer tokens per frame may allow for longer contexts and better results.
Second, our method often cuts on action between two unrelated videos where motion is similar.
This is common in projects having shots where one would expect a cut on action to occur, but there is no footage that would create a coherent cut on action.
Fig.~\ref{fig:bad-cut} shows an example of this failure mode.
Third, the sequencing and trimming of video is but one task in the endeavor of film post-production.
Other tasks include editing audio, color grading, and shot transitions different from cuts.
Though our current model is conditioned on audio, it cannot directly modify it in any way.
This limitation leaves some film idioms out of reach of our model, most notably L-cuts and J-cuts~\cite{hockrow2014}.
Finally, the video editing task is one where a number of condition signals would help in producing better results.
Some examples of useful condition signals are screenplays, storyboards, or explicit style directions.
Our current version of this model only receives conditioning from audio and constraints in the video collection through our inference algorithm, which is effective but limited.
We envision future versions of this model to be capable of receiving more complex conditioning signals and style.

Ethical concerns include potential impact on labor and possible malicious use. Just as spell-checkers and auto-complete did not replace the role of the writer, our goal is not to automate away the role of the human film editor.
Our model handles the low- and mid-level tasks of video editing, such as pacing, continuity, and action matching.
This frees the human editor to focus on high-level aspects of editing: story-crafting, emotional arcs, and meaning.
As for potentially malicious use, our model is limited to splicing existing footage with cinematic-quality cuts. Thus, unlike models that synthesize new visual content, FilmGPT is unlikely to fabricate videos or hallucinate events that did not occur. However, as with any editing tool, the system could be misused to selectively recontextualize footage in misleading ways through omission or rearrangement. We emphasize that FilmGPT is designed as an assistive, human-in-the-loop system: it proposes candidate edits, but editorial intent, factual integrity, and ethical responsibility remain with the user.

\section{Conclusion}
In this paper, we show a simple and effective approach to computational video montage using autoregressive sequence modeling. Our approach achieves a new state of the art for shot sequence ordering and out-performs previous approaches in quality of edited results according to user studies. Taken together, our contributions show that large-scale sequence modeling provides a practical and effective way to learn and apply the ``grammar'' of film editing directly from data. More broadly, we see our work as part of a longer trajectory in graphics and vision: just as computational photography has dramatically improved the quality of everyday images without automating artistic intent, our work opens up a new avenue of human-computer co-creation for making existing video more watchable. In a world already saturated with video, the challenge is not to generate more, but to rescue from obscurity what has already been captured.

\bibliographystyle{ACM-Reference-Format}
\bibliography{main}

\end{document}